\documentclass{article} % For LaTeX2e
\usepackage{iclr2025_conference,times}

\usepackage{amsmath,amsfonts,bm}

\def\eqref#1{equation~\ref{#1}}
\def\1{\bm{1}}

\def\rvs{{\mathbf{s}}}

\def\rvw{{\mathbf{w}}}
\def\rvx{{\mathbf{x}}}
\def\rvy{{\mathbf{y}}}

\DeclareMathAlphabet{\mathsfit}{\encodingdefault}{\sfdefault}{m}{sl}
\SetMathAlphabet{\mathsfit}{bold}{\encodingdefault}{\sfdefault}{bx}{n}

\def\gD{{\mathcal{D}}}

\def\gL{{\mathcal{L}}}

\def\gN{{\mathcal{N}}}

\newcommand{\E}{\mathbb{E}}

\newcommand{\R}{\mathbb{R}}

\newcommand{\KL}{D_{\mathrm{KL}}}

\usepackage{hyperref}
\usepackage{url}
\usepackage{amssymb}

\usepackage{enumitem}

\usepackage{url}            % simple URL typesetting
\usepackage{booktabs}       % professional-quality tables% 
\usepackage{multirow}
\usepackage{siunitx} 
\usepackage{cancel}
\usepackage{tablefootnote}

\usepackage{amsmath,amsfonts,amssymb}       % blackboard math symbols
\usepackage{nicefrac}       % compact symbols for 1/2, etc.
\usepackage{microtype}      % microtypography
\usepackage{float}
\usepackage{graphicx}

\usepackage{booktabs} % For better table lines
\usepackage{array}    % For table wrapping and better column spacing

\usepackage{algorithm,algorithmic}

\usepackage{wrapfig}

\usepackage{microtype}
\usepackage{graphicx}
\usepackage{subcaption}
\usepackage{booktabs} 
\usepackage{chngcntr}

\usepackage{enumitem}

\usepackage{amsmath}
\usepackage{amssymb}
\usepackage{mathtools}
\usepackage{amsthm}
\usepackage{textcomp}
\usepackage{boldline}
\usepackage{colortbl} 
\usepackage{makecell}
\usepackage{caption}
\usepackage{fontenc} 
\usepackage{inputenc}
\definecolor{lightgray}{gray}{0.9}
\definecolor{midgray}{gray}{0.7}

\usepackage{newtxmath}

\usepackage[createShortEnv,conf={no link to proof, text link},commandRef=autoref]{proof-at-the-end}

\newcommand{\spacehack}[1]{\relax}

\makeatletter

\renewcommand{\appendixautorefname}{\S\@gobble}
\renewcommand{\sectionautorefname}{\S\@gobble}
\renewcommand{\subsectionautorefname}{\S\@gobble}
\renewcommand{\subsubsectionautorefname}{\S\@gobble}

\makeatother

\newcommand{\post}{^{\rm post}}
\newcommand{\LRTB}{\gL_{\rm RTB}}

\newcommand{\ie}{\textit{i.e.}}

\newcommand{\std}[1]{\textcolor{black}{\scriptsize{$\pm #1$}}}

\title{Solving Bayesian inverse problems with diffusion priors and off-policy RL}

\author{%
    Luca Scimeca\textsuperscript{*}\\Mila, Universit\'e de Montr\'eal
    \And
    Siddarth Venkatraman\textsuperscript{*}\\Mila, Universit\'e de Montr\'eal
    \And
    Moksh Jain\textsuperscript{*}\\Mila, Universit\'e de Montr\'eal
    \And
    Minsu Kim\textsuperscript{*}\\Mila, Universit\'e de Montr\'eal\\KAIST
    \And
    Marcin Sendera\textsuperscript{*}\\Mila, Universit\'e de Montr\'eal\\Jagiellonian University
    \And
    Mohsin Hasan\\Mila, Universit\'e de Montr\'eal
    \And
    Luke Rowe\\Mila, Universit\'e de Montr\'eal
    \And
    Sarthak Mittal\\Mila, Universit\'e de Montr\'eal
    \And
    Pablo Lemos\\Mila, Universit\'e de Montr\'eal\\Ciela Institute\\Dreamfold
    \And
    Emmanuel Bengio\\Recursion
    \And
    Alexandre Adam\\Mila, Universit\'e de Montr\'eal\\Ciela Institute
    \And
    Jarrid Rector-Brooks\\Mila, Universit\'e de Montr\'eal\\Dreamfold
    \And
    Yashar Hezaveh\\Mila, Universit\'e de Montr\'eal\\Ciela Institute
    \And
    Laurence Perreault-Levasseur\\Mila, Universit\'e de Montr\'eal\\Ciela Institute
    \And
    Yoshua Bengio\\Mila, Universit\'e de Montr\'eal\\CIFAR
    \And
    Glen Berseth\\Mila, Universit\'e de Montr\'eal\\CIFAR
    \And
    Nikolay Malkin\\Mila, Universit\'e de Montr\'eal\\University of Edinburgh
    \AND
    \tt
$\left\{\text{\begin{minipage}{3.5in}\centering luca.scimeca,siddarth.venkatraman,moksh.jain,\\minsu.kim,marcin.sendera,\dots,nikolay.malkin\end{minipage}}\right\}$@mila.quebec
}

\iclrfinalcopy % Uncomment for camera-ready version, but NOT for submission.
\begin{document}

\maketitle

\begin{abstract}
This paper presents a practical application of Relative Trajectory Balance (RTB), a recently introduced off-policy reinforcement learning (RL) objective that can asymptotically solve Bayesian inverse problems optimally. We extend the original work by using RTB to train conditional diffusion model posteriors from pretrained unconditional priors for challenging linear and non-linear inverse problems in vision, and science. We use the objective alongside techniques such as off-policy backtracking exploration to improve training. Importantly, our results show that existing training-free diffusion posterior methods struggle to perform effective posterior inference in latent space due to inherent biases. 

\end{abstract}

\section{Introduction}
\label{sec:intro}

\looseness=-1
While deep learning has seen rapid advancements, scientific discovery, particularly in high-dimensional and multimodal contexts, remains a significant challenge. Many scientific problems, such as inverse protein design and gravitational lensing, can be framed as Bayesian inverse problems~\citep{kaipio2006statistical,idier2013bayesian,dashti2013bayesian,latz2020well} due to the inherent uncertainties, often introduced by imperfections in scientific instruments. Traditionally, the scientific community has approached Bayesian inference using methods like Markov chain Monte Carlo (MCMC), or more efficient alternatives like variational inference~\citep{mackay2003information}, Hamiltonian Monte Carlo~\citep{duane1987hybrid,neal2012mcmc}, and Langevin dynamics~\citep{besag1994comments, roberts1996exponential, roberts1998optimal}. However, these methods become impractical when applied to complex, real-world scenarios.

Recently, diffusion models~\citep{sohl2015diffusion,ho2020ddpm,song2021score} have emerged as a promising approach for tackling Bayesian inverse problems. Diffusion models \cite{sohl2015diffusion,ho2020ddpm,song2021score} are a powerful class of hierarchical generative models, used to model complex distributions over a varied range of objects, including images \cite{nichol2021improvedddpm,dhariwal2021diffusion,rombach2021high}, text \citep{austin2021structured,dieleman2022continuous,li2022diffusion,han-etal-2023-ssd,gulrajani2024likelihood,lou2023discrete}, actions in reinforcement learning, \cite{janner2022diffuser,wang2023diffusion,kang2024efficient}, proteins and more besides. In each of these domains, downstream problems require sampling product distributions, where a pretrained diffusion model serves as a prior $p(\rvx)$ that is multiplied by an auxiliary constraint $r(\rvx)$. For example, if $p(\rvx)$ is a prior over images defined by a diffusion model, and $r(\rvx)=p(c\mid\rvx)$ is the likelihood that an image $\rvx$ belongs to class $c$, then class-conditional image generation requires sampling from the Bayesian posterior $p(\rvx\mid c)\propto p(\rvx)p(c\mid\rvx)$.

The hierarchical nature of the generative process in diffusion models, which generate samples from $p(\rvx)$ by a deep chain of stochastic transformations, makes exact sampling from posteriors $p(\rvx)r(\rvx)$ under a black-box function $r(\rvx)$ intractable. Common solutions to this problem involve inference techniques based on linear approximations \cite{song2022solving,kawar2021snips,kadkhodaie2021solving,chung2023diffusion} or stochastic optimization \cite{graikos2022diffusion,mardani2024variational}. Others estimate the `guidance' term -- the difference in drift functions between the diffusion models sampling the prior and posterior -- by training a classifier on noised data \cite{dhariwal2021diffusion}, but when such data is not available, one must resort to approximations or Monte Carlo estimates \citep{song2023loss,dou2024diffusion,cardoso2024montecarlo}, which are challenging to scale to high-dimensional problems. Reinforcement learning methods that have recently been proposed for this problem \cite{black2024training, fan2023reinforcement} are biased and prone to mode collapse \cite{venkatraman2024amortizing}.

Recently, \cite{venkatraman2024amortizing} introduced an asymptotically unbiased objective for finetuning a diffusion prior to sample from the Bayesian posterior. The objective was named \textit{relative trajectory balance} (RTB) due to its relationship with the trajectory balance objective \citep{malkin2022trajectory}, as they both arise from the generative flow network perspective of diffusion models \citep{lahlou2023theory,zhang2022unifying}. RTB is an asymptotically unbiased objective for finetuning a diffusion prior to sample from the Bayesian posterior, and has been presented as an alternative to existing on-policy, policy gradient-based methods~\cite{black2024training, fan2023reinforcement} due to its off-policy training capabilities. However, unlike other methods, RTB has not been thoroughly evaluated on complex, real-world challenges, raising questions about its scalability to high-dimensional problems and its feasibility in such settings.

In this paper, we demonstrate the effectiveness of RTB through its application to intractable linear and nonlinear Bayesian inverse problems in vision and the scientific application of gravitational lensing. We challenge the uncertainties over prior work by providing empirical evidence that RTB, when combined with off-policy adaptation techniques (e.g., as introduced in~\citep{zhang2021path,sendera2024improvedoffpolicytrainingdiffusion}), significantly improves scalability to complex scientific discovery problems. We also extend RTB to train conditional diffusion posteriors from unconditional priors, and add experiments combining this objective with other state-of-the-art techniques (e.g. DPS~\cite{chung2023diffusion} or FPS~\cite{song2023loss}). Our findings demonstrate that RTB is not only a viable off-the-shelf objective for Bayesian inverse problems in scientific domains but also offer comprehensive guidance for practitioners on scaling RTB to high-dimensional settings.

To summarize, our contributions are as follows: 
\begin{itemize}[left=0pt]
\item We demonstrate that RTB can effectively address a wide range of complex, linear and non-linear Bayesian inverse problems in vision. W
\item also provide empirical evidence that specific off-policy adaptation techniques enhance RTB, offering practical insights for real-world applications.
\item We extend RTB to train conditional diffusion posteriors from unconditional priors.
\item We extend RTB to integrate additional state of the art tecniques (e.g. DPS \& FPS)
\item We conduct a extensive benchmarks of prior methods, empirically revealing the limitations of existing training-free methods.
\end{itemize}

\section{Solving Bayesian inverse problems with relative trajectory balance}

\paragraph{Inverse problems.} A typical inverse problem is the following: We are interested in recovering some quantity $\rvx \sim p(\rvx)$. However, in the process of measurement, the quantity of interest is perturbed by some noise, or instrumental systematic effect. The new observation $\rvy \sim p(\rvy)$ contains information about the observation of interest, but it has been distorted by the experiment. Furthermore, we assume (as it is often the case) that we have a good enough understanding of our instrumentation, to be able to compute $p(\rvy | \rvx)$, \ie, if we assume a true underlying $\rvx$, we know how likely it is to recover our observation. What we are interested in, however, is $p(\rvx \mid \rvy)$, \ie, given our observation, how likely is a given value of $\rvx$. 

Inverse problems such as these are very common in various scientific disciplines, but can be extremely ill-posed, particularly if the noise is complex and non-linear, and if the quantities of interest are high-dimensional. Traditional methods, such as Markov-Chain Monte Carlo, quickly become unusable on complex problems, such as the ones we illustrate in Section  \ref{sec:experiments} of this paper. Advances in generative modelling~\citep{song2021score} have made diffusion models suitable for learning rich and expressive priors from data for inverse problems~\citep{adam2022posterior}. 

\paragraph{Summary of setting.} 
% We review the background and setup for diffusion models in \autoref{sec:background}. In short, a diffusion model (in discretized time) describes a Markovian generative process $\rvx_0\rightarrow\rvx_{\Delta t}\rightarrow\dots\rightarrow\rvx_1$ by means of a drift function $u_\theta$, representing the drift term in an It\^o stochastic differential equation (SDE). The initial, or noise, distribution $p(\rvx_0)$ (typically a standard normal) and the process parametrized by $\theta$ induce a terminal distribution $p_\theta(\rvx_1)$, which is used as a prior over $\rvx_1$ in the inference problems we consider. 

% \subsection{Diffusion models as hierarchical generative models}
% \label{sec:diffusion_prior}
%%GB This section should be moved before this section and be called the background section. This background inside the method section is very confusing.

A denoising diffusion model generates data $\rvx_1$ by a Markovian generative process:
\begin{equation}\label{eq:diffusion_model}
    \text{\it (noise)}\quad \rvx_0\rightarrow\rvx_{\Delta t}\rightarrow\rvx_{2\Delta t}\rightarrow\ldots\rightarrow\rvx_1=\rvx\quad\text{\it (data)},
\end{equation}
where $\Delta t=\frac1T$ and $T$ is the number of discretization steps.\footnote{\looseness=-1 The time indexing suggestive of an SDE discretization is used for consistency with the diffusion samplers literature \cite{zhang2021path,sendera2024diffusion}. The indexing $\rvx_T\rightarrow\rvx_{T-1}\rightarrow\dots\rightarrow\rvx_0$ is often used for diffusion models trained from data.} The initial distribution $p(\rvx_0)$ is fixed (typically to $\gN(\boldsymbol{0},\boldsymbol{I})$) and the transition from $\rvx_{t-1}$ to $\rvx_t$ is modeled as a Gaussian perturbation with time-dependent variance:
\begin{equation}\label{eq:diffusion_transition}
    p(\rvx_{t+\Delta t}\mid\rvx_t)=\gN(\rvx_{t+\Delta t}\mid\rvx_t+u_t(\rvx_t)\Delta t,\sigma_t^2\Delta t\boldsymbol{I}).
\end{equation}
The scaling of the mean and variance by $\Delta t$ is insubstantial for fixed $T$, but ensures that the diffusion process is well-defined in the limit $T\to\infty$ assuming regularity conditions on $u_t$. The process given by (\ref{eq:diffusion_model}, \ref{eq:diffusion_transition}) is then identical to Euler-Maruyama integration of the stochastic differential equation (SDE) $d\rvx_t=u_t(\rvx_t)\,dt+\sigma_t\,d\rvw_t$.
% \begin{equation}\label{eq:diffusion_sde}
%     d\rvx_t=u_t(\rvx_t)\,dt+\sigma_t\,d\rvw_t.
% \end{equation}

The likelihood of a denoising trajectory $\rvx_0\rightarrow\rvx_{\Delta t}\rightarrow\dots\rightarrow\rvx_1$ factors as
\begin{equation}\label{eq:diffusion_likelihood}
    p(\rvx_0,\rvx_{\Delta t},\dots,\rvx_1)=p(\rvx_0)\prod_{i=1}^T p(\rvx_{i\Delta t}\mid\rvx_{(i-1)\Delta t})
\end{equation}
and defines a marginal density over the data space:
\begin{equation}\label{eq:diffusion_marginal}
    p(\rvx_1)=\int p(\rvx_0,\rvx_{\Delta t},\dots,\rvx_1)\,d\rvx_0\,d\rvx_{\Delta t}\ldots d\rvx_{1-\Delta t}.
\end{equation}
A reverse-time process, $\rvx_1\rightarrow\rvx_{1-\Delta t}\rightarrow\dots\rightarrow\rvx_0$, with densities $q$, can be defined analogously, and similarly defines a conditional density over trajectories:
\begin{equation}\label{eq:diffusion_likelihood_reverse}
    q(\rvx_0,\rvx_{\Delta t},\dots,\rvx_{1-\Delta t}\mid \rvx_1)=\prod_{i=1}^T q(\rvx_{(i-1)\Delta t}\mid\rvx_{i\Delta t}).
\end{equation}
In the training of diffusion models, as discussed below, the process $q$ is typically fixed to a simple distribution (usually a discretized Ornstein-Uhlenbeck process), and the result of training is that $p$ and $q$ are close as distributions over trajectories. In this work, we consider diffusino model priors.

\paragraph{Intractable inference under a diffusion prior.} Consider a diffusion model $p_\theta$, defining a marginal density $p_\theta(\rvx_1)$, and a positive constraint function $r:\R^d\to\R_{>0}$. We are interested in training a diffusion model $p_\phi\post$, with drift function $u_\phi\post$, that would sample the product distribution $p\post(\rvx_1)\propto p_\theta(\rvx_1)r(\rvx_1)$. If $r(\rvx_1)=p(\rvy\mid\rvx_1)$ is a conditional distribution over another variable $\rvy$, then $p\post$ is the Bayesian posterior $p_\theta(\rvx_1\mid\rvy)$.

\looseness=-1
Because samples from $p\post(\rvx_1)$ are not assumed to be available, one cannot directly train $p$ using the forward KL objective. Nor can one directly apply objectives for distribution-matching training, such as those that enforce the trajectory balance (TB) constraint, since the marginal $p_\theta(\rvx_1)$ is not available. However, \cite{venkatraman2024amortizing} makes the observation that an alternate constraint relates the denoising process which samples from the posterior to the one which samples from the prior, and proposes an objective as a function of the vector $\phi$ that parametrizes the posterior diffusion model and the scalar $Z_\phi$ (parametrized via $\log Z_\phi$ for numerical stability) as follows:
\begin{equation}\label{eq:rtb_objective}
    \LRTB(\rvx_0\rightarrow\rvx_{\Delta t}\rightarrow\dots\rightarrow\rvx_1;\phi)
    :=
    \left(\log\frac
        {Z_\phi\cdot p_\phi\post(\rvx_0,\rvx_{\Delta t},\dots,\rvx_1)}{r(\rvx_1)p_\theta(\rvx_0,\rvx_{\Delta t},\dots,\rvx_1)}
    \right)^2.
\end{equation}
When all prior assumptions hold, optimizing this objective to 0 for all trajectories ensures that $p\post_{\phi}(\rvx_1) \propto p_{\theta}(\rvx_1)r(\rvx_1)$. 

Notably, the gradient of this objective with respect to $\phi$ does not require differentiation (backpropagation) into the sampling process that produced a trajectory $\rvx_0\rightarrow\dots\rightarrow\rvx_1$. This offers two advantages over on-policy simulation-based methods: (1) the ability to optimize $\LRTB$ as an off-policy objective, \ie, sampling trajectories for training from a distribution different from $p_\phi\post$ itself, as discussed further in \autoref{sec:training}; (2) backpropagating only to a subset of the summands in (\ref{eq:rtb_objective}), when computing and storing gradients for all steps in the trajectory is prohibitive for large diffusion models. We discuss further details about the training and parametrization in \autoref{sec:training}.

\section{Training, parametrization, and conditioning} 
\label{sec:training}

\paragraph{Training and exploration.}
%Training with RTB requires choosing for which trajectories to evaluate and take gradient steps on the loss, akin to the choice of behavior policy in RL
The choice of which trajectories we use to take gradient steps with the RTB loss can have a large impact on sample efficiency. In \emph{on-policy} training, we use the current policy $p_\phi\post$ to generate trajectories $\tau =(\rvx_0\rightarrow\ldots\rightarrow\rvx_1)$, evaluate the reward $\log r(\rvx_1)$ and the likelihood of $\tau$ under $p_\theta$, and a gradient updates on $\phi$  to minimize $\LRTB(\tau;\phi)$.
%%%$ (\ref{eq:rtb_objective}) on the trajectory.  %$\mathcal{L}(\phi) = \Bigl(\log \frac{Z_{\phi}p_{\phi}\post(\tau)}{r(\rvx_1)p_{\theta}(\tau)}\Bigr)^2$ for these trajectories with any optimizer such as Adam~\citep{kingma2014adam}. 

However, on-policy training may be insufficient to discover the modes of the posterior distribution.
%%GB: Why? This I don't understand. This sound like we are saying that off-policy training is better to fund the modes. What is the off line data is insufficient?
In this case, we can perform \emph{off-policy} exploration to ensure mode coverage. For instance, given samples $\rvx_1$ that have high density under the target distribution, we can sample \emph{noising} trajectories $\rvx_1 \leftarrow \rvx_{1-\Delta t} \leftarrow \ldots \leftarrow \rvx_0$ starting from these samples and use such trajectories for training. Another effective off-policy training technique uses replay buffers. We expect the flexibility of mixing on-policy training with off-policy exploration to be a strength of RTB over on-policy RL methods, as was shown for distribution-matching training of diffusion models in \cite{sendera2024diffusion}.

% TODO B&F maybe, pending Moksh experiment (Kolya - duplicate todo, find best place for it)

\paragraph{Conditional constraints and amortization.}
%%GB: Is this section used somewhere else in the paper? I'm not sure how this is connected or useful. A reference from the experiments section is needed for this discussion.
We extend the RTB objective to amortize conditional posterior inference from an unconditional diffusion prior. If the constraints depend on other variables $\rvy$ -- for example, $r(\rvx_1;\rvy)=p(\rvy\mid\rvx_1)$ -- then the posterior drift $u_\phi\post$ can be conditioned on $\rvy$ and the learned scalar $\log Z_\phi$ replaced by a model taking $\rvy$ as input. In this case, $Z_{\phi}$ is thus a function of the conditioning variable $Z_{\phi}(\rvs)$. For continuous variables $\rvs$ or if the number of categories for discrete $\rvs$ are large, we can parametrize $\phi$ as a neural network. Such conditioning achieves amortized inference and allows generalization to new $\rvy$ not seen in training. Similarly, all of the preceding discussion easily generalizes to \emph{priors} that are conditioned on some context variable, yielding:
\begin{equation}\label{eq:cond_rtb_objective}
    \LRTB(\rvx_0\rightarrow\rvx_{\Delta t}\rightarrow\dots\rightarrow\rvx_1;\rvs,\phi)
    :=
    \left(\log\frac
        {Z_\phi(\rvs)\cdot p_\phi\post(\rvx_0,\rvx_{\Delta t},\dots,\rvx_1 \mid \rvs)}{r(\rvx_1, \rvs)p_\theta(\rvx_0,\rvx_{\Delta t},\dots,\rvx_1 \mid \rvs)}
    \right)^2
\end{equation}

\paragraph{Efficient parametrization and Langevin inductive bias.} Because the deep features learned by the prior model $u_\theta$ are expected to be useful in expressing the posterior drift $u_\phi\post$, we can choose to initialize $u_\phi\post$ as a copy of $u_\theta$ and to fine-tune it, possibly in a parameter-efficient way (as described in each section of \autoref{sec:experiments}). This choice is inspired by the method of amortizing inference in large language models by fine-tuning a prior model to sample an intractable posterior \cite{hu2023amortizing}.

Furthermore, if the constraint $r(\rvx_1)$ is differentiable, we can impose an inductive bias on the posterior drift similar to the one introduced for diffusion samplers of unnormalized target densities in \cite{zhang2021path} and shown to be useful for off-policy methods in \cite{sendera2024diffusion}. namely, we write
\begin{equation}\label{eq:langevin}
    u_{\phi}\post(\rvx_t, t) = \text{NN}_1(\rvx_t, t; \phi) + \text{NN}_2(\rvx_t, t, \phi) \nabla_{\rvx_t}\log r(\rvx_t),
\end{equation}
where $\text{NN}_1$ and $\text{NN}_2$ are neural networks outputting a vector and a scalar, respectively. This parametrization allows the constraint to provide a signal to guide the sampler at intermediate steps. % comment that this resembles also classifier guidance?

\paragraph{Stabilizing the loss.} We propose two simple design choices for stabilizing RTB training. First, the loss in (\ref{eq:rtb_objective}) can be replaced by the empirical \emph{variance} over a minibatch of the quantity inside the square, which removes dependence on $\log Z_\phi$ and is especially useful in conditional settings, consistent with the findings of \cite{sendera2024diffusion}. This amounts to a relative variant of the VarGrad objective~\citep{richter2020vargrad}. Second, we employ loss clipping: to reduce sensitivity to an imperfectly fit prior model, we do not perform updates on trajectories where the loss is close to 0.

\subsection{Off-policy adaptation techniques}

Similar to methods used in GFlowNets, RTB benefits from the powerful advantages of off-policy learning. This flexibility allows us to choose any behavior policy over trajectories \( \tau \), denoted as \( P(\tau) \), independent of the current diffusion sampling process. To leverage this advantage, we employ three off-policy techniques to enhance the performance of RTB for posterior inference.

\textbf{Backtracking exploration with replay buffer}

Using a replay buffer \( \mathcal{D} = \{(x_1, r(x_1))\} \) with a prioritizing distribution \( P(x_1; \mathcal{D}) \) can prevent catastrophic forgetting of the sampler, such as mode dropping. By leveraging the off-policy property of RTB, we utilize a behavior policy defined as
\[
P_{\beta}(\tau) = P_B(\tau \mid x_1) \, P(x_1; \mathcal{D}),
\]
where we train the RTB objective over $\tau \sim P_{\beta}(\tau)$.

\section{Experiments} \label{sec:experiments}

In this section, we demonstrate the wide applicability of RTB to sample from complex image posteriors with diffusion priors, and highlight important shortcomings of current methods.

% The explanation for the inverse problems can also be moved to the supplementary section shuold we need space

\subsubsection*{Linear inverse problems}
\paragraph{Inpainting}

Inpainting is a classical inverse problem where the goal is to reconstruct missing or occluded parts of an image \citep{chung2023diffusion}. Let $\rvx$ represent the original image, and the forward operator $A(\rvx) = P \rvx$, where $P$ is a masking matrix that zeros out the missing pixels, representing the incomplete observation. The measurement $\rvy$ is the partially observed image, and is subject to noise of scale $\sigma$, so $\rvy = P \mathbf{x} + \gN(0,\sigma^2 \mathbf{I})$. The task is to infer the posterior distribution $p(\mathbf{x} \mid \mathbf{y})$. We consider two types of inpainting tasks: random inpainting and box inpainting. In random inpainting, the mask $P$ is applied randomly to a set of pixels, removing a random subset of the image. Box inpainting, instead, is a variant where a large rectangular region of the image is removed \citep{kadkhodaie2021solving}. 
We use RTB to fine-tune a score-based prior $p_\theta(\mathbf{x})$ into a posterior $p_\theta(\rvx) p(\rvy \mid \rvx)$ with likelihood $p(\rvy \mid \rvx) \propto \exp\left(-\frac{\|P \rvx - \rvy\|^2}{2 \sigma^2}\right)$.

\subsubsection*{Non-linear inverse problems}
We consider two classic non-linear inverse problems, i.e. Fourier phase retrieval, and nonlinear deblur.

\paragraph{Fourier phase retrieval}

Fourier phase retrieval is a classical inverse problem in which the objective is to recover a signal from its Fourier magnitude \citep{fienup1987phase}. The challenge lies in the loss of the phase information during the measurement process, making the inverse problem highly ill-posed and non-unique~\citep{chung2023diffusion}. Let $\rvx$ represent the original signal, and the forward operator $A(\rvx) = |\mathcal{F}(\rvx)|$ denotes the magnitude of the Fourier transform. The measurement $\rvy$ is the observed Fourier magnitude corrupted by noise of scale $\sigma$, so $\rvy= |\mathcal{F}(\mathbf{x})| + \gN(0,\sigma^2\mathbf{I})$.
The inverse problem is to infer the posterior distribution $p(\mathbf{x}\mid\mathbf{y})$. We use RTB to fine-tune a score-based prior $p_\theta(\mathbf{x})$ into an unbiased posterior $p_\theta(\rvx) p(\rvy \mid \rvx)$ with likelihood $p( \rvy \mid \rvx) \propto \exp\left(-\frac{\||\rvy - \mathcal{F}(\rvx)| \|^2}{2\sigma^2}\right)$, for sample $\rvx$ and reference measurement $\mathbf{y}$, and where $\sigma$ controls the temperature of the likelihood. 

\paragraph{Nonlinear deblur}

For nonlinear deblur, the objective is to recover a clean image from its blurry observation. The forward model generally involves complex, nonlinear, transformations, such as temporal integration of sharp images through a nonlinear camera response function. We leverage the neural network-based forward model $A(\rvx)$ as described by \citet{nah2017deep}. We can thus describe the measurement $\rvy$ as $\rvy = A(\rvx) + \gN(0, \sigma^2 \mathbf{I})$, where $\rvx$ represent the original sharp image, and the forward operator $A(\rvx)$ encapsulate the nonlinear blurring process. The use RTB to fine-tune a posterior $p(\mathbf{x} \mid \mathbf{y})$, considering the likelihood $p(\rvy \mid \rvx) \propto \exp\left(-\frac{\|A(\rvx) - \rvy\|^2}{2 \sigma^2}\right)$. 

\begin{figure}
\centering
% \vspace*{-1em}
\includegraphics[width=.9\linewidth]{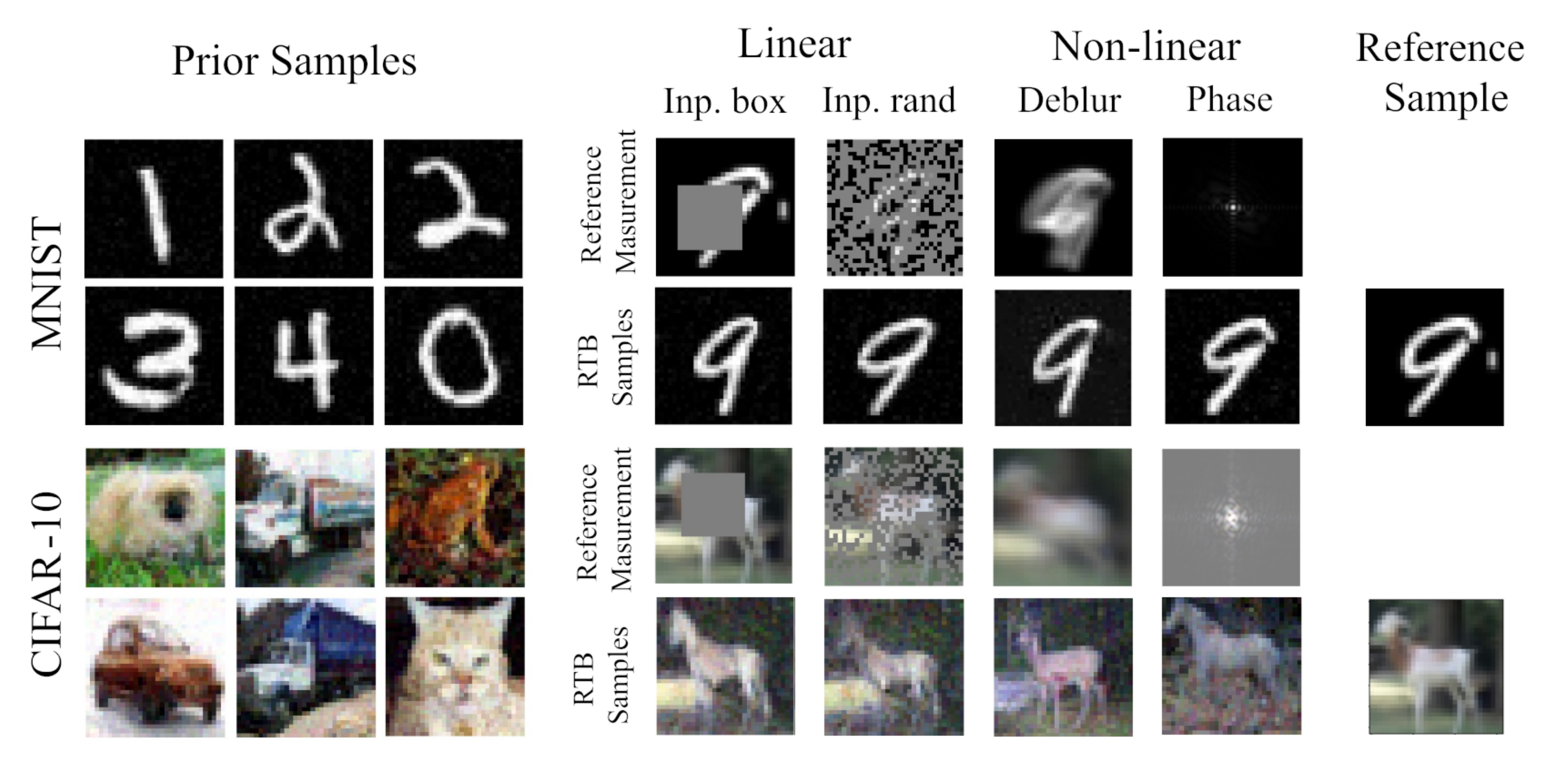}
\caption{Samples from RTB fine-tuned diffusion posteriors.} 
\label{fig:vision_inv_problems_main}
\end{figure}
\subsubsection*{Results}

We consider diffusion score-based priors with MNIST and CIFAR-10, and re-implement several methods previously showing state-of-the art results in inverse problems in vision, including DPS \cite{chung2023diffusion}, FPS \cite{song2023loss} and FPS-SMC \cite{dou2024diffusion}.

\subsubsection*{Unconditional Posteriors}
In the first set of experiments we finetune unconditional diffusion priors into unconditional posteriors for a measurement $\rvy$. For each method and dataset we run experiments on inverse problems defined above, and report average performance over 10 randomly sampled measurements retrieved from the validation set of the respective datasets. We report in Table \autoref{tab:unconditional_vision} the mean log reward ($\E[\log r(\rvx)]$) of the generated samples, their LPIPS score and the log-partition function $\log Z$, computed with 5000 posterior samples, and averaged across 10 measurements.

We observe classifier guidance (CF)-based methods to achieve high $\E[\log r(\rvx)]$ and LPIPS values at the expense of drifting from the true posterior to model (low $\log Z$ values). On the other hand, RTB reaches competitive rewards while maintaining significantly higher values of $\log Z$, thus getting closer to the true posterior.

\begin{table}[t]
% \vspace*{-1em}
\caption{\looseness=-1 Results for linear and nonlinear inverse problems on pretrained standard diffusion models. We report the mean and standard error of each metric across MNIST and CIFAR-10 datasets.}
\label{tab:unconditional_vision}
\centering
\resizebox{1\linewidth}{!}{
                                    \begin{tabular}{@{}lcccccccccccc@{}}\toprule
                  \multicolumn{13}{c}{\textbf{Linear Inverse Problems }} \\\midrule
                  Task $\rightarrow$ & \multicolumn{6}{c}{Inpainting (box)} & \multicolumn{6}{c}{Inpainting (random)} \\\cmidrule(lr){2-7}\cmidrule(lr){8-13}
                  Dataset $\rightarrow$ & \multicolumn{3}{c}{MNIST} & \multicolumn{3}{c}{CIFAR-10} 
                          & \multicolumn{3}{c}{MNIST} & \multicolumn{3}{c}{CIFAR-10} 
                          \\\cmidrule(lr){2-4}\cmidrule(lr){5-7}\cmidrule(lr){8-10}\cmidrule(lr){11-13} 
                  Algorithm $\downarrow$ Metric $\rightarrow$ 
                          & $\E[\log r(\rvx)]$ ($\uparrow$)  & $\log Z$ ($\uparrow$)  & LPIPS ($\downarrow$)  
                          & $\E[\log r(\rvx)]$ ($\uparrow$)  & $\log Z$ ($\uparrow$)  & LPIPS ($\downarrow$)  
                          & $\E[\log r(\rvx)]$ ($\uparrow$)  & $\log Z$ ($\uparrow$)  & LPIPS ($\downarrow$)  
                          & $\E[\log r(\rvx)]$ ($\uparrow$)  & $\log Z$ ($\uparrow$)  & LPIPS ($\downarrow$)  
                          \\\midrule
                  DPS                  & $-0.759$ \std{ 0.054 } & $-2067.326$ \std{ 2.15e+02 } & $0.107$ \std{ 0.011 } & $-0.680$ \std{ 0.083 } & $-51810.868$ \std{ 2.07e+03 } & $0.083$ \std{ 0.009 } & $-1.360$ \std{ 0.017 } & $-2945.032$ \std{ 1.67e+02 } & $0.136$ \std{ 0.010 } & $-0.577$ \std{ 0.047 } & $-29942.605$ \std{ 9.83e+02 } & $0.052$ \std{ 0.005 } \\
FPS                  & $-1.010$ \std{ 0.140 } & $-1200.852$ \std{ 1.20e+02 } & $0.141$ \std{ 0.010 } & $-0.778$ \std{ 0.035 } & $-44381.465$ \std{ 1.44e+03 } & $0.107$ \std{ 0.010 } & $-1.121$ \std{ 0.072 } & $-2089.714$ \std{ 2.75e+02 } & $0.169$ \std{ 0.010 } & $-1.009$ \std{ 0.021 } & $-42189.443$ \std{ 9.54e+02 } & $0.077$ \std{ 0.007 } \\
FPS-SMC              & $-0.902$ \std{ 0.153 } & $-1205.810$ \std{ 1.28e+02 } & $0.128$ \std{ 0.010 } & $-0.801$ \std{ 0.053 } & $-42424.722$ \std{ 1.52e+03 } & $0.108$ \std{ 0.011 } & $-1.053$ \std{ 0.067 } & $-1891.618$ \std{ 1.87e+02 } & $0.163$ \std{ 0.010 } & $-1.103$ \std{ 0.024 } & $-40025.465$ \std{ 9.39e+02 } & $0.077$ \std{ 0.007 } \\
\textbf{RTB (ours)}           & $-3.133$ \std{ 0.205 } & $-18.122$ \std{ 1.49e+00 } & $0.181$ \std{ 0.013 } & $-11.022$ \std{ 0.670 } & $-36.567$ \std{ 3.79e+00 } & $0.493$ \std{ 0.034 } & $-2.940$ \std{ 0.144 } & $-19.189$ \std{ 1.35e+00 } & $0.172$ \std{ 0.012 } & $-12.152$ \std{ 1.34e+00 } & $-25.276$ \std{ 2.72e+00 } & $0.547$ \std{ 0.039 } \\
\midrule
                   \multicolumn{13}{c}{\textbf{Nonlinear Inverse Problems}} \\\midrule
                   Task $\rightarrow$ & \multicolumn{6}{c}{Phase Retrieval} & \multicolumn{6}{c}{Nonlinear Deblur} \\\cmidrule(lr){2-7}\cmidrule(lr){8-13}
                   Dataset $\rightarrow$ & \multicolumn{3}{c}{MNIST} & \multicolumn{3}{c}{CIFAR-10} 
                           & \multicolumn{3}{c}{MNIST} & \multicolumn{3}{c}{CIFAR-10} 
                           \\\cmidrule(lr){2-4}\cmidrule(lr){5-7}\cmidrule(lr){8-10}\cmidrule(lr){11-13} 
                   Algorithm $\downarrow$ Metric $\rightarrow$ 
                           & $\E[\log r(\rvx)]$ ($\uparrow$)  & $\log Z$ ($\uparrow$)  & LPIPS ($\downarrow$)  
                           & $\E[\log r(\rvx)]$ ($\uparrow$)  & $\log Z$ ($\uparrow$)  & LPIPS ($\downarrow$)  
                           & $\E[\log r(\rvx)]$ ($\uparrow$)  & $\log Z$ ($\uparrow$)  & LPIPS ($\downarrow$)  
                           & $\E[\log r(\rvx)]$ ($\uparrow$)  & $\log Z$ ($\uparrow$)  & LPIPS ($\downarrow$)  
                           \\\midrule
                   DPS                  & $-1.722$ \std{ 0.235 } & $-1625.894$ \std{ 1.12e+02 } & $0.184$ \std{ 0.023 } & $-2.782$ \std{ 0.246 } & $-20885.408$ \std{ 2.82e+03 } & $0.560$ \std{ 0.020 } & $-1.698$ \std{ 0.088 } & $-1536.865$ \std{ 5.28e+01 } & $0.145$ \std{ 0.009 } & $-2.672$ \std{ 0.100 } & $-33036.342$ \std{ 1.08e+03 } & $0.191$ \std{ 0.012 } \\
FPS                  & $-2.123$ \std{ 0.242 } & $-1860.426$ \std{ 2.16e+02 } & $0.212$ \std{ 0.021 } & $-2.910$ \std{ 0.167 } & $-43860.793$ \std{ 6.10e+03 } & $0.567$ \std{ 0.020 } & $-1.760$ \std{ 0.092 } & $-1890.115$ \std{ 1.17e+02 } & $0.150$ \std{ 0.010 } & $-3.190$ \std{ 0.125 } & $-60519.596$ \std{ 2.33e+03 } & $0.218$ \std{ 0.013 } \\
FPS-SMC              & $-2.058$ \std{ 0.241 } & $-1394.984$ \std{ 7.22e+01 } & $0.205$ \std{ 0.021 } & $-2.865$ \std{ 0.164 } & $-36354.029$ \std{ 5.07e+03 } & $0.566$ \std{ 0.021 } & $-1.585$ \std{ 0.070 } & $-1842.228$ \std{ 8.26e+01 } & $0.120$ \std{ 0.009 } & $-21.893$ \std{ 1.54e+00 } & $-49.469$ \std{ 2.51e+00 } & $0.654$ \std{ 0.010 } \\
\textbf{RTB (ours)}           & $-3.600$ \std{ 0.209 } & $-17.986$ \std{ 1.26e+00 } & $0.184$ \std{ 0.021 } & $-9.284$ \std{ 0.439 } & $-27.573$ \std{ 2.76e+00 } & $0.566$ \std{ 0.033 } & $-3.286$ \std{ 0.156 } & $-15.573$ \std{ 1.30e+00 } & $0.181$ \std{ 0.012 } & $-6.964$ \std{ 0.289 } & $-36.034$ \std{ 2.17e+00 } & $0.440$ \std{ 0.026 } \\
\bottomrule\end{tabular}
}
\end{table}

\subsubsection*{Conditional Posteriors}

In the second set of experiments, we fine-tune unconditional diffusion priors into conditional posteriors. We follow the formulation in \autoref{eq:cond_rtb_objective}, and condition the generation of the posterior model by the corresponding measurement. For each method and dataset we run experiments on all inverse problems previously defined. We report in Table \autoref{tab:unconditional_vision} the mean log reward ($\E[\log r(\rvx)]$) of the generated samples, their FID score and the log-partition function $\log Z$, all computed with 10000 posterior samples and respective measurements.

Similarly to before, classifier guidance (CF)-based methods to achieve high $\E[\log r(\rvx)]$ and low FID values at the expense of drifting from the true posterior to model (low $\log Z$ values). On the other hand, RTB reaches competitive rewards while maintaining significantly higher values of $\log Z$. Importantly, perform experiments whereby DPS and  DPS + CLA drifts are added to the RTB objective following a similar formulation to \autoref{eq:langevin}. We obverse increased rewards when blending these methods, mitigating the original pitifals of exceptionally low $\log Z$ values.

\begin{table}[t]
% \vspace*{-1em}
\caption{\looseness=-1 Conditional Diffusion results for linear and nonlinear inverse problems on pretrained standard diffusion models. We report the mean and standard error of each metric across MNIST and CIFAR-10 datasets.}
\centering
\resizebox{1\linewidth}{!}{
\begin{tabular}{@{}lcccccccccccc@{}}\toprule
                  \multicolumn{13}{c}{\textbf{Linear Inverse Problems }} \\\midrule
                  Task $\rightarrow$ & \multicolumn{6}{c}{Inpainting (box)} & \multicolumn{6}{c}{Inpainting (random)} \\\cmidrule(lr){2-7}\cmidrule(lr){8-13}
                  Dataset $\rightarrow$ & \multicolumn{3}{c}{MNIST} & \multicolumn{3}{c}{CIFAR-10} 
                          & \multicolumn{3}{c}{MNIST} & \multicolumn{3}{c}{CIFAR-10} 
                          \\\cmidrule(lr){2-4}\cmidrule(lr){5-7}\cmidrule(lr){8-10}\cmidrule(lr){11-13} 
                  Algorithm $\downarrow$ Metric $\rightarrow$ 
                          & $\E[\log r(\rvx)]$ ($\uparrow$)  & $\log Z$ ($\uparrow$)  & FID ($\downarrow$)  
                          & $\E[\log r(\rvx)]$ ($\uparrow$)  & $\log Z$ ($\uparrow$)  & FID ($\downarrow$)  
                          & $\E[\log r(\rvx)]$ ($\uparrow$)  & $\log Z$ ($\uparrow$)  & FID ($\downarrow$)  
                          & $\E[\log r(\rvx)]$ ($\uparrow$)  & $\log Z$ ($\uparrow$)  & FID ($\downarrow$)  
                          \\\midrule
                  DPS                  & $-0.509$ \std{ 0.004 } & $-484.746$ & $0.450$ & $-0.279$ \std{ 0.001 } & $-15309.502$ & $0.065$ & $-0.383$ \std{ 0.002 } & $-870.558$ & $1.317$ & $-0.220$ \std{ 0.001 } & $-13771.698$ & $0.253$ \\
DPS+CLA              & $-0.507$ \std{ 0.004 } & $-550.389$ & $0.446$ & $-0.273$ \std{ 0.001 } & $-14988.690$ & $0.066$ & $-0.379$ \std{ 0.002 } & $-800.751$ & $1.312$ & $-0.220$ \std{ 0.001 } & $-13287.473$ & $0.253$ \\
FPS                  & $-0.784$ \std{ 0.005 } & $-1036.824$ & $0.941$ & $-0.519$ \std{ 0.000 } & $-14281.504$ & $0.104$ & $-0.610$ \std{ 0.003 } & $-844.831$ & $1.391$ & $-0.495$ \std{ 0.000 } & $-12541.325$ & $0.304$ \\
FPS-SMC              & $-0.671$ \std{ 0.005 } & $-1031.968$ & $0.478$ & $-0.289$ \std{ 0.001 } & $-12216.635$ & $0.073$ & $-0.470$ \std{ 0.004 } & $-744.221$ & $1.356$ & $-0.267$ \std{ 0.001 } & $-14835.040$ & $0.295$ \\
\textbf{RTB (ours)}           & $-0.704$ \std{ 0.001 } & $-655.195$ & $1.515$ & $-1.394$ \std{ 0.001 } & $-2445.463$ & $0.804$ & $-1.048$ \std{ 0.001 } & $-628.810$ & $1.397$ & $-1.976$ \std{ 0.002 } & $-2459.556$ & $0.774$ \\
\textbf{RTB+DPS (ours)}       & $-0.579$ \std{ 0.001 } & $-801.660$ & $1.532$ & $-0.491$ \std{ 0.001 } & $-7600.880$ & $0.191$ & $-0.358$ \std{ 0.001 } & $-1722.438$ & $1.485$ & $-0.330$ \std{ 0.001 } & $-9852.971$ & $0.674$ \\
\textbf{RTB+DPS+CLA (ours)}   & $-0.527$ \std{ 0.001 } & $-807.689$ & $1.457$ & $-0.350$ \std{ 0.001 } & $-7704.646$ & $0.482$ & $-0.340$ \std{ 0.001 } & $-1982.161$ & $1.539$ & $-0.314$ \std{ 0.001 } & $-11071.914$ & $0.681$ \\
\midrule
                   \multicolumn{13}{c}{\textbf{Nonlinear Inverse Problems}} \\\midrule
                   Task $\rightarrow$ & \multicolumn{6}{c}{Phase Retrieval} & \multicolumn{6}{c}{Nonlinear Deblur} \\\cmidrule(lr){2-7}\cmidrule(lr){8-13}
                   Dataset $\rightarrow$ & \multicolumn{3}{c}{MNIST} & \multicolumn{3}{c}{CIFAR-10} 
                           & \multicolumn{3}{c}{MNIST} & \multicolumn{3}{c}{CIFAR-10} 
                           \\\cmidrule(lr){2-4}\cmidrule(lr){5-7}\cmidrule(lr){8-10}\cmidrule(lr){11-13} 
                   Algorithm $\downarrow$ Metric $\rightarrow$ 
                           & $\E[\log r(\rvx)]$ ($\uparrow$)  & $\log Z$ ($\uparrow$)  & FID ($\downarrow$)  
                           & $\E[\log r(\rvx)]$ ($\uparrow$)  & $\log Z$ ($\uparrow$)  & FID ($\downarrow$)  
                           & $\E[\log r(\rvx)]$ ($\uparrow$)  & $\log Z$ ($\uparrow$)  & FID ($\downarrow$)  
                           & $\E[\log r(\rvx)]$ ($\uparrow$)  & $\log Z$ ($\uparrow$)  & FID ($\downarrow$)  
                           \\\midrule
                   DPS                  & $-1.249$ \std{ 0.009 } & $-1630.394$ & $1.237$ & $-2.402$ \std{ 0.011 } & $-12885.045$ & $0.541$ & $-1.675$ \std{ 0.004 } & $-986.008$ & $1.402$ & $-2.214$ \std{ 0.005 } & $-10682.691$ & $0.442$ \\
DPS+CLA              & $-1.272$ \std{ 0.009 } & $-1372.473$ & $1.241$ & $-2.414$ \std{ 0.011 } & $-14362.470$ & $0.531$ & $-1.667$ \std{ 0.004 } & $-780.577$ & $1.413$ & $-2.211$ \std{ 0.005 } & $-11472.098$ & $0.441$ \\
FPS                  & $-1.715$ \std{ 0.011 } & $-1029.060$ & $1.424$ & $-2.842$ \std{ 0.011 } & $-12805.804$ & $0.682$ & $-1.754$ \std{ 0.005 } & $-815.970$ & $1.386$ & $-2.353$ \std{ 0.006 } & $-9779.317$ & $0.495$ \\
FPS-SMC              & $-1.668$ \std{ 0.011 } & $-1117.023$ & $1.374$ & $-2.796$ \std{ 0.011 } & $-11583.854$ & $0.657$ & $-1.948$ \std{ 0.006 } & $-324.646$ & $1.479$ & $-6.934$ \std{ 0.013 } & $-276.732$ & $0.849$ \\
\textbf{RTB (ours)}           & $-2.860$ \std{ 0.010 } & $8960.105$ & $1.440$ & $-4.551$ \std{ 0.010 } & $-968.128$ & $1.642$ & $-2.406$ \std{ 0.004 } & $-204.788$ & $1.464$ & $-2.827$ \std{ 0.005 } & $-1912.678$ & $1.018$ \\
\textbf{RTB+DPS (ours)}       & $-6.705$ \std{ 0.020 } & $-58.640$ & $1.548$ & $-2.622$ \std{ 0.010 } & $-6290.827$ & $1.305$ & $-1.896$ \std{ 0.004 } & $-638.202$ & $1.413$ & $-2.253$ \std{ 0.005 } & $-4747.092$ & $0.873$ \\
\textbf{RTB+DPS+CLA (ours)}   & $-2.906$ \std{ 0.009 } & $-1321.135$ & $1.382$ & $-2.354$ \std{ 0.009 } & $-11500.541$ & $1.087$ & $-1.849$ \std{ 0.003 } & $-791.639$ & $1.305$ & $-2.701$ \std{ 0.004 } & $-3525.784$ & $1.635$ \\
\bottomrule\end{tabular}
}
\end{table}

\subsection{Gravitational lensing}
In general relativity, light travels along the shortest paths in a spacetime curved by the mass of objects \citep{einstein1916foundation}, with greater masses inducing larger curvature. An interesting inverse problem involves the inference of the undistorted images of distant astronomical sources whose images have been gravitationally lensed by the gravity of intervening structures \citep{einstein1911graviton}. 
In the case of strong lensing, for example when the background source and the foreground lens are both almost perfectly aligned galaxies, multiple images of the background source are formed and heavy distortions such as rings or arcs are induced. In this problem, the parameters of interest are the undistorted pixel values of the background source $x$, given an observed distorted image $y$.  This problem is then linear, since the distortions can be encoded in a lensing matrix $A$ (which we assume to be known):  $\mathbf{y} = A\mathbf{x} + \mathbf{\epsilon}$, with $\mathbf{\epsilon} \sim \mathcal{N}(\mathbf{0}, \sigma^2\mathbf{I})$ a small Gaussian observational noise.
%\begin{equation}
%    \mathbf{y} = A\mathbf{x} + \mathbf{\epsilon}
%\end{equation}
The Bayesian inverse problem of interest is the inference of the posterior distribution over source images given the lensed observation, that is $p(\mathbf{x} \mid \mathbf{y})$. We use the Probes dataset~\cite{Probes2022}, containing telescope images of undistorted galaxies in the local Universe, to train a score based prior over source images $p_{\theta}(\mathbf{x})$. Drawing unbiased samples from the posterior $p(\mathbf{x} \mid \mathbf{y}) \propto p_{\theta}(\mathbf{x})\mathcal{N}(\mathbf{y}; A\mathbf{x}, \sigma^2\mathbf{I})$ is quite difficult, especially if the distribution is very peaky with small $\sigma$. RTB allows us to train an asymptotically unbiased posterior sampler.

\begin{figure}
\centering
% \vspace{-10pt}
\includegraphics[width=\linewidth]{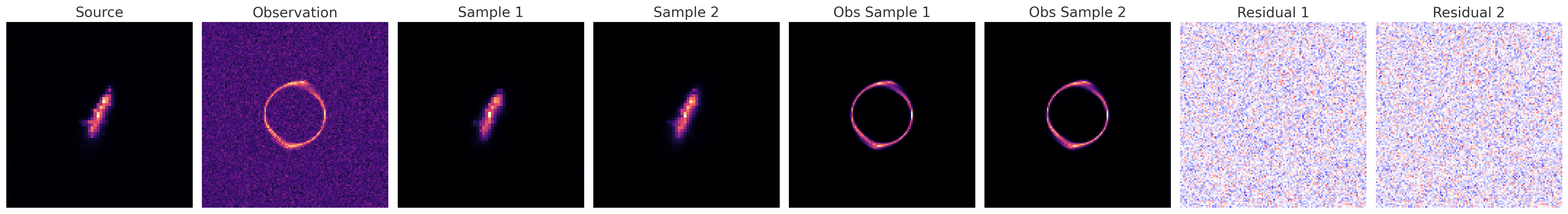}
\caption{Lensing problem RTB samples. Plotted are ground truth source, observation, samples from RTB posterior, their mean observation after forward model (without observation white noise), and the residual between posterior mean observations and ground truth observation.} 
\label{fig:astro_samples}
\vspace{-12pt}
\end{figure}

We use RTB to finetune the prior model to this posterior, and compare against a biased training-free diffusion posterior inference baseline \citep{adam2022posterior} that previous work has used for this gravitational lensing inverse problem. This method uses a convolved likelihood approximation (CLA) $p_t(\mathbf{y} \mid \mathbf{x}) \approx \mathcal{N}(\mathbf{y} \mid A\mathbf{x}, (\sigma^2 + \sigma^2(t))\mathbf{I})$. For RTB we use 300 diffusion steps for sampling, but for CLA we require 2000 steps to obtain reasonable samples. We fix $\sigma = 0.05$ for our experiments. We report metrics comparing these approaches in \autoref{tab:lensing}, and illustrative samples in \autoref{fig:astro_samples}. We found RTB to be a bit unstable while training, likely because of the peaky reward function. About 30\% of runs, the policy diverged irrecoverably. For the sake of highlighting the advantages of unbiased posterior sampling, the metrics computed in \autoref{tab:lensing} excluded diverged runs. For this problem, we only used on-policy samples, and we expect off-policy tricks such as replay buffers to help stabilize training.

\begin{table}[]
\vspace*{-1em}
    \centering
    \begin{minipage}{0.57\textwidth}
    \captionof{table}{Comparison between RTB and CLA for the lensing problem. We compare mean likelihood $\log p(\mathbf{y} \mid \mathbf{x})$, and lower bound on the log-partition function $\log Z$. Metrics are computed with 50 posterior samples, and averaged across 3 runs.\label{tab:lensing}}
    \end{minipage}\hfill
    \begin{minipage}{0.39\textwidth}
\vspace*{-2em}
\resizebox{\textwidth}{!}{\begin{tabular}{lccc}
        \toprule
         Algorithm & $\log p(\mathbf{y} \mid \mathbf{x}) (\uparrow)$& $\log Z (\uparrow)$\\
         \midrule
         CLA & $-8216.02$ &  $-12514.67$\\
         RTB & $-8367.9 $& $-8676.85$\\ %& $-17907.39$\\
         \bottomrule
    \end{tabular}}
    \end{minipage}
\end{table}

\section{Discussion}

Our study has demonstrated the potential of relative trajectory balance (RTB) as an effective framework for solving Bayesian inverse problems with diffusion models. By systematically evaluating RTB across a variety of inverse problems, including vision-based reconstructions and gravitational lensing, we provide comprehensive empirical evidence supporting its scalability and flexibility.

\subsection{Comparison with Existing Methods}

RTB exhibits several advantages over existing approaches such as diffusion posterior sampling (DPS) \citep{chung2023diffusion} and function-space posterior sampling (FPS) \citep{song2023loss}. While these methods provide practical solutions for Bayesian inverse problems, they often tend to shift the distribution of samples away from the true posterior, leading to artificially high likelihood values at the cost of reduced diversity. As an asymptotically unbiased objective, RTB mitigates these issues while enabling flexible off-policy training strategies. Our results indicate that RTB achieves competitive likelihood scores while preserving a closer match to the true posterior, as evidenced by higher log-partition function ($\log Z$) values.

\subsection{Extensions and Future Directions}

Despite its strong empirical performance, RTB presents certain challenges that warrant further exploration.

\paragraph{Off-Policy Stabilization:} While we leveraged replay buffers to prevent mode collapse, stability during training remains a challenge in some settings. Future work could explore improved adaptive strategies for selecting trajectories and incorporating uncertainty estimation during training.

\paragraph{Integration with Other State-of-the-Art Methods:} Our experiments have shown that RTB can benefit from hybrid approaches, such as combining DPS-based constraints with RTB optimization. Further research could explore novel ways to blend RTB with other advanced generative modeling techniques to enhance robustness and scalability.

\paragraph{Application to Broader Scientific Domains:} We have primarily focused on inverse problems in vision and gravitational lensing. However, RTB has the potential to generalize to other scientific fields, such as climate modeling, computational biology, and medical imaging. Future studies should investigate its applicability in these domains, particularly in handling multimodal constraints and heterogeneous data.

\section{Conclusion}

In this work, we demonstrated the effectiveness of off-policy RL fine-tuning via the RTB objective for asymptotically unbiased posterior inference for diffusion models. We applied RTB to challenging linear and non-linear Bayesian inverse problems, demonstrating its effectiveness in inverse imaging and gravitational lensing. The ability to seamlessly integrate off-policy training and other classifier guidance techniques, as well as the extension of the objective for conditional posteriors, allows for RTB to be leveraged in more complex and critical domains at scale. Extending RTB to other important scientific applications, such as inverse protein design would be a promising direction for future research.

\subsubsection*{Acknowledgments}
The authors acknowledge funding from CIFAR, NSERC, IVADO, UNIQUE, FACS Acuit\'e, NRC AI4Discovery, Samsung, and Recursion.

The research was enabled in part by computational resources provided by the Digital Research
Alliance of Canada (\url{https://alliancecan.ca}), Mila (\url{https://mila.quebec}), and
NVIDIA.

\bibliography{iclr2025_conference}
\bibliographystyle{iclr2025_conference}
\newpage
\clearpage
\appendix
\counterwithin{figure}{section}
\counterwithin{table}{section}

\section{Background and setup}\label{sec:background}

\paragraph{Diffusion model training as divergence minimization.}
Diffusion models parametrize the drift $u_t(\rvx_t)$ in (\autoref{eq:diffusion_transition}) as a neural network $u(\rvx_t,t;\theta)$ with parameters $\theta$ and taking $\rvx_t$ and $t$ as input. We denote the distributions over trajectories induced by (\autoref{eq:diffusion_likelihood}, \autoref{eq:diffusion_marginal}) by $p_\theta$ to show their dependence on the parameter. 

In the most common setting, diffusion models are trained to maximize the likelihood of a dataset. In the notation above, this corresponds to assuming $q(\rvx_1)$ is fixed to an empirical measure (with the points of a training dataset $\gD$ assumed to be i.i.d.\ samples from $q(\rvx_1)$). Training minimizes with respect to $\theta$ the divergence between the processes $q$ and $p_\theta$:
\begin{align}\label{eq:diffusion_kl_objective}
    &\KL(q(\rvx_0,\rvx_{\Delta t},\dots,\rvx_1)\,\|\,p_\theta(\rvx_0,\rvx_{\Delta t},\dots,\rvx_1))\\
    &=\KL(q(\rvx_1)\,\|\,p_\theta(\rvx_1))+\mathbb{E}_{\rvx_1\sim q(\rvx_1)}\KL(q(\rvx_0,\rvx_{\Delta t},\dots,\rvx_{1-\Delta t}\mid \rvx_1)\,\|\,p_\theta(\rvx_0,\rvx_{\Delta t},\dots,\rvx_{1-\Delta t}\mid \rvx_1))
    \nonumber\\&\geq \KL(q(\rvx_1)\,\|\,p_\theta(\rvx_1))
    =\mathbb{E}_{\rvx_1\sim q(\rvx_1)}[-\log p_\theta(\rvx_1)]+\text{\rm const}.\nonumber
\end{align}
where the inequality -- an instance of the data processing inequality for the KL divergence -- shows that minimizing the divergence between distributions over trajectories is equivalent to maximizing a lower bound on the data log-likelihood under the model $p_\theta$. 

As shown in \cite{song2021maximum}, minimization of the KL in (\autoref{eq:diffusion_kl_objective}) is essentially equivalent to the traditional approach to training diffusion models via denoising score matching \cite{vincent2011connection,sohl2015diffusion,ho2020ddpm}. Such training exploits that for typical choices of the noising process $q$, the optimal $u_t(\rvx_t)$ can be expressed in terms of the Stein score of $q(\rvx_1)$ convolved with a Gaussian, allowing an efficient stochastic regression objective for $u_t$. For full generality of our exposition for arbitrary iterative generative processes, we prefer to think of (\autoref{eq:diffusion_kl_objective}) as the primal objective and denoising score matching as an efficient means of minimizing it.

% Minsu
\subsection{ Diffusion GFlowNets}

Generative Flow Networks (GFlowNets) aim to sample from a distribution proportional to an unnormalized density, $p(x) \propto r(x)$, through a sequential decision-making process. Diffusion GFlowNets are a family of GFlowNets that model discretized reverse stochastic differential equation (SDE) trajectories,
\begin{equation}
\tau = \left( x_0 \rightarrow x_{\Delta t} \rightarrow x_{2\Delta t} \rightarrow \ldots \rightarrow x_1 \right),
\end{equation}
where $x_0 = (\boldsymbol{0}, t = 0)$ is the initial state. Here, $\Delta t = \frac{1}{T}$, where $T$ is the number of discrete time steps.

The forward policy $P_F(x_{t + \Delta t} \mid x_t; \theta)$ is defined to model the mean of a Gaussian kernel, expressed as:
\begin{equation}
P_F(x_{t + \Delta t} \mid x_t; \theta) = \mathcal{N}\left( x_{t + \Delta t}; \, x_t + u(x_t, t; \theta) \Delta t, \, \sigma(t)^2 \Delta t \, \mathbb{I} \right).
\end{equation}
Here, \( u(x_t, t; \theta) \) is the learnable score with parameter $\theta$, and \( \sigma(t) \) represents the standard deviation at time \( t \). The term \( \mathbb{I} \) denotes the identity matrix, ensuring that the covariance matrix is isotropic.

The backward policy $P_B(x_{t - \Delta t} \mid x_t)$ is defined as a discretized Brownian bridge with a noise rate $\sigma$:
\begin{equation}
P_B(x_{t - \Delta t} \mid x_t) = \mathcal{N}\left( x_{t - \Delta t}; \, \frac{t - \Delta t}{t} x_t, \, \frac{t - \Delta t}{t} \sigma^2 \Delta t \, \mathbb{I} \right).
\end{equation}
The forward and backward policies defined over complete trajectories are expressed as:
\begin{equation}
P_F(\tau; \theta) = \prod_{i=0}^{T-1} P_F\left( x_{(i+1)\Delta t} \mid x_{i\Delta t}; \theta \right), \quad 
P_B(\tau; x_1) = \prod_{i=0}^{T-1} P_B\left( x_{i\Delta t} \mid x_{(i+1)\Delta t} \right).
\end{equation}

\begin{equation}
\mathcal{L}_{\text{TB}}(\tau; \theta, f(x_1)) = \log \left( \frac{Z_{\theta} \, P_F(\tau; \theta)}{f(x_1) \, r_{\text{target}}(x_1) \, P_B(\tau; x_1)} \right),
\end{equation}

where \( Z_{\theta} \) is a learnable constant representing the partition function, \( f(x_1) \) is a weighting function for the density, and \( r_{\text{target}}(x_1) \) is the accessible unnormalized true density.

By ensuring that \( \mathcal{L}_{\text{TB}}(\tau; \theta, f(x_1)) = 0 \) for all trajectories \( \tau \), we guarantee an optimal amortized sampler over the weighted distribution. This condition ensures that the distribution satisfies \( p(x_1) \propto f(x_1) \, r(x_1) \). Specifically, when the weighting function is set to \( f(x_1) = 1 \), the target distribution simplifies to \( p(x_1) \propto r(x_1) \).

\section{Relative trajectory balance (RTB)}

\subsection{Method}

In diffusion GFlowNets, the Trajectory Balance (TB) objective is employed to perform amortized inference, aiming to approximate the distribution \( p(x_1) \propto r(x_1) \). In contrast, Relative Trajectory Balance (RTB) focuses on amortized posterior inference over the prior distribution. Specifically, RTB defines the posterior as

\[
p^{\text{post}}(x_1; \theta) \propto p^{\text{prior}}(x_1) \, r(x_1),
\]

where \( p^{\text{prior}}(x_1) \) is the prior distribution (e.g., a learned diffusion model) and \( r(x_1) \) represents the likelihood. The product \( p^{\text{prior}}(x_1) \, r(x_1) \) serves as the target unnormalized density for amortized inference of $p^{\text{post}}(x_1; \theta)$.

RTB is TB that has weighted reward $p^{\text{prior}}(x_1) \, r(x_1)$, where the weight is prior distribution $p^{\text{prior}}(x_1)$:

\begin{align}
\mathcal{L}_{\text{TB}}(\tau; \theta, p^{\text{prior}}(x_1)) &= \log \left( \frac{Z_{\theta} \, P^{\text{post}}_F(\tau; \theta)}{p^{\text{prior}}(x_1) \, r(x_1) \,  P^{\text{post}}_B(\tau; x_1)} \right) \\
&= \log \left( \frac{Z_{\theta} \, P^{\text{post}}_F(\tau; \theta) \cancel{P^{\text{prior}}_B(\tau; x_1)}}{r(x_1) \, P_F^{\text{prior}}(\tau) \, \cancel{P^{\text{post}}_B(\tau; x_1)}} \right) \\
&= \log \left( \frac{Z_{\theta} \, P^{\text{post}}_F(\tau; \theta)}{r(x_1) \, P_F^{\text{prior}}(\tau)} \right) \\
&= \mathcal{L}_{\text{RTB}}(\tau; \theta).
\end{align}

The cancellation arises from the fact that \( P^{\text{post}}_B(\tau; x_1) = P^{\text{prior}}_B(\tau; x_1) \), since we assumed the backward policy to be a fixed Brownian bridge with \(\sigma\) noise.

% \section{Related works}
% \color{red}TODO\color{black}

% \newpage

% \appendix

% \section{Additional results}

% \section{Experiment details}

\end{document}